\pdfoutput=1
\documentclass[11pt]{article}

\usepackage{acl}

\usepackage{times}
\usepackage{latexsym}

\usepackage[T1]{fontenc}
\usepackage[utf8]{inputenc}

\usepackage{microtype}

\usepackage{graphicx}
\usepackage{algorithm}
\usepackage[noend]{algpseudocode}

\title{Robust Product Classification with Instance-Dependent Noise}

\author{Huy Nguyen \\
  Amazon.com, Inc. \\
  Seattle, Washington, USA \\
  \texttt{nguynnq@amazon.com} \\\And
  Devashish Khatwani \\
  Amazon.com, Inc. \\
  Vancouver, British Columbia, Canada \\
  \texttt{khatwad@amazon.com} \\}

\begin{document}
\maketitle
\begin{abstract}
Noisy labels in large E-commerce product data (i.e., product items are placed into incorrect categories) are a critical issue for product categorization task because they are unavoidable, non-trivial to remove and degrade prediction performance significantly. Training a product title classification model which is robust to noisy labels in the data is very important to make product classification applications more practical. In this paper, we study the impact of instance-dependent noise to performance of product title classification by comparing our data denoising algorithm and different noise-resistance training algorithms which were designed to prevent a classifier model from over-fitting to noise. We develop a simple yet effective Deep Neural Network for product title classification to use as a base classifier. Along with recent methods of stimulating instance-dependent noise, we propose a novel noise stimulation algorithm based on product title similarity. Our experiments cover multiple datasets, various noise methods and different training solutions. Results uncover the limit of classification task when noise rate is not negligible and data distribution is highly skewed. 
\end{abstract}

\section{Introduction}
\label{sec:intro}

Product classification is a quintessential E-commerce machine learning problem in which product items are placed into their respective categories. With recent advancements of Deep Learning, various unimodal (i.e., text only) and multimodal (e.g., text and image) models have been developed to predict larger numbers of items and categories with better accuracy \cite{gao_deep_2020, chen_multimodal_2021, brinkmann_improving_2021}. However, one of the fundamental assumptions behind such models is the availability of large and high-quality labeled datasets. Access to such datasets is usually costly or infeasible in some settings. Large product datasets usually suffer from annotation errors, i.e., products are assigned to incorrect categories, partially due to complex category structure, confusing categories and similar titles. The problem of noisy labels is even more severe when product category distribution is highly imbalanced with heavy-tail \cite{shen_large-scale_2012, das_large-scale_2016}. Therefore, a text classifier which is robust to noisy labels present in training data is critical for high-performing product classification applications.

While machine learning in the presence of label noise has been studied for decades, most of prior studies experimented in computer vision domain \cite{gu_realistic_2021, song_learning_2022}, and only a few research was conducted in text classification \cite{jindal_effective_2019, garg_towards_2021}. Without an annotated dataset with manually-identified label noise, classical approaches for label noise stimulation assume class-conditional noise (CCN) where the probability of an item having label corrupted depends on the original and noisy labels. With this assumption, all products of ``Men's Watches'' category have the sample probability to be assigned ``Women's Watches'' label. This is not generally correct. For instance, product titles having phrase ``men's watches'' are less likely mis-labeled. Recent research addresses more general label noise, i.e., instance-dependent noise (IDN), that an item is mis-labeled with a probability depending on its original label and features.

In this paper, we present a comprehensive study on improving product title classification in the presence of IDN. We develop a simple yet effective Deep Neural Network for text classification and show that our model performs well on different product title datasets ranging from small to medium sizes, balanced to skewed distributions, and tens to over a hundred categories. To generate noisy labels for experiments, our first contribution is an IDN stimulation algorithm which flips an item's label based on its similarity to items of other categories. Noisy label data generated by our method is compared with prior IDN stimulation methods for their impact to model accuracy degradation. To make the model robust to label noise, our second contribution is a data augmentation method that reduces noise rate and thus improves model's accuracy. We compare three state-of-the-art Deep Neural Network training algorithms to train a classifier on data with label noise generated by different methods. From experimental results we discuss lessons learned for product title classification in production. To the best of our knowledge, this work is the first time that noise-resistance model training is studied in E-commerce domain, which is our third contribution.

\section{Related Work}
\label{sec:work}

Automatic product categorization has been well studied to address its challenges including large number of items and categories, and hierarchical categories structure \cite{gao_deep_2020, chen_multimodal_2021, brinkmann_improving_2021}. The large-scale nature of product data leads to a critical issue of noisy labels. For example, an E-commerce website reported that 15\% of product listings by sellers have incorrect labels \cite{shen_large-scale_2012}. \citet{das_large-scale_2016} attempted to use a latent topic model to help manually inspect noisy categories and remove incorrect samples. Our current study focuses on fully automated methods for data denoising and noise-resistance training to prevent models from over-fitting to noisy samples.

Training Deep Neural Networks (DNN) with noisy labels is challenging because DNN's large learning capacity make them highly susceptible to over-fitting to noise \cite{arpit_closer_2017, zhang_understanding_2021}. Early work stacked DNN with layers to model noise-transition matrix assuming class-conditional noise, i.e., noisy label $\hat{y}$ only depends on true label $y$ but not on the input $x$ \cite{jindal_learning_2016, patrini_making_2017}. Because noise transition matrix can be difficult to learn or not feasible in real-world settings, other directions targeted to selecting clean samples in each mini-batch and use them to update DNN's parameters \cite{jiang_mentornet_2018, malach_decoupling_2017}. Among those, CoTeaching \cite{han_co-teaching_2018} and CoTeaching$^+$ \cite{yu_how_2019} showed the effectiveness of cross-training two networks simultaneously in that each network sends selective samples for the other to learn. A more realistic assumption of noisy labels is instance-dependent noise (IDN) in which probability of noisy label $\hat{y}$ depends on true label $y$ and input $x$ \cite{chen_beyond_2021}. Among state-of-the-art work on IDN, Self-Evolution Average Label -- SEAL \cite{chen_beyond_2021} and Progressive Label Correction -- PLC \cite{zhang_learning_2021} are representatives of label refurbishment \cite{song_learning_2022} that uses softmax output to assign soft labels to training instances. We compare SEAL, PLC and CoTeaching$^+$ on training a product title classifier with label noise.

\section{Datasets}
\label{sec:data}

In this study, we employ 6 public datasets for product classification. While some datasets have multimodal inputs, e.g., product titles, descriptions, images, we use only product title  inputs and leave other fields for a future work. This restriction may prevent us from achieving the best possible performance by incorporating other information-rich inputs \cite{chen_multimodal_2021}. However, our main motivation is to evaluate noise-resistance training approaches. For each dataset, we filter-out category labels with less than 10 samples, then apply stratified random sampling to split 10\% for testing and 90\% for training. We leave a study of few-shot learning for product title classification for future work. Hyper-parameters of models and training algorithms are fine-tuned within training sets when needed. In experiments with noisy labels, only training samples have label corrupted while testing sets are unchanged. This assures a realistic evaluation that model accuracies are measured against ground-truth disregarding how the model was trained. To measure skewness of data label distribution, we calculate KL-divergence from the actual category distribution to uniform distribution. Data statistics are shown in Table~\ref{data}.

\begin{itemize}
	\item Flipkart\footnote{www.kaggle.com/PromptCloudHQ/flipkart-products}: the original set contains nearly 20,000 samples but over 200 category labels are unqualified for modeling (e.g., those either have too few samples or are considered as Brand Name). Therefore we use 19,666 samples of top 28 categories.
	\item WDC dataset is WDC-25 Gold Standard for Product Categorization \cite{primpeli_wdc_2019}. We remove items with category label ``\emph{not-found}'' and keep 23,597 samples with 24 class labels.
	\item Retail dataset has 46,228 training samples with item titles, descriptions, images and category labels placed into 21 categories \cite{elayanithottathil_retail_2021}. We do not use their test data which does not have category labels.
	\item Pricerunner, Shopmania, Skroutz datasets\footnote{www.kaggle.com/lakritidis/product-classification-and-categorization} were collected from three online electronic stores and product comparison platforms \cite{akritidis_effective_2018, akritidis_self-verifying_2020}.
\end{itemize}

As shown in Table~\ref{data}, datasets Flipkart, Shopmania and Skroutz are highly imbalanced with KL-divergence greater than 1. Each of these datasets has major classes with thousands of samples and minor classes with tens of samples. WDC dataset is moderately skewed having 24 classes with number of samples ranging from 10 to 4,753. Retail and Pricerunner sets are the most balanced with KL-divergence close to zero. Retail dataset has roughly 2,200 samples per class while Pricerunner has class samples in range (2000, 6000).

\begin{table}
  \caption{Summary of product title datasets}
  \label{data}
  \centering
  \begin{tabular}{lrrrr}
    \hline
    Dataset     & \#cls     & \#train & \#test & KL \\
    \hline
    Flipkart & 28  & 17,682 & 1,984 & 1.04 \\
    WDC & 24  & 21,225 & 2,372 & 0.34 \\
    Retail & 21  & 41,586 & 4,642 & 0.00 \\
    Pricerunner & 10  & 31,773 & 3,538 & 0.03 \\
    Shopmania & 147 & 282,095  & 31,437 & 1.49 \\
    Skroutz & 12  & 214,346 & 23,824 & 1.10 \\
    \hline
  \end{tabular}
\end{table}

\section{Base Model for Product Title Categorization}
\label{sec:base}

\begin{figure*}
  \caption{LSTM-CNNs architecture for product title classifier}
  \label{fig:network}
  \centering
  \includegraphics[width=0.7\textwidth]{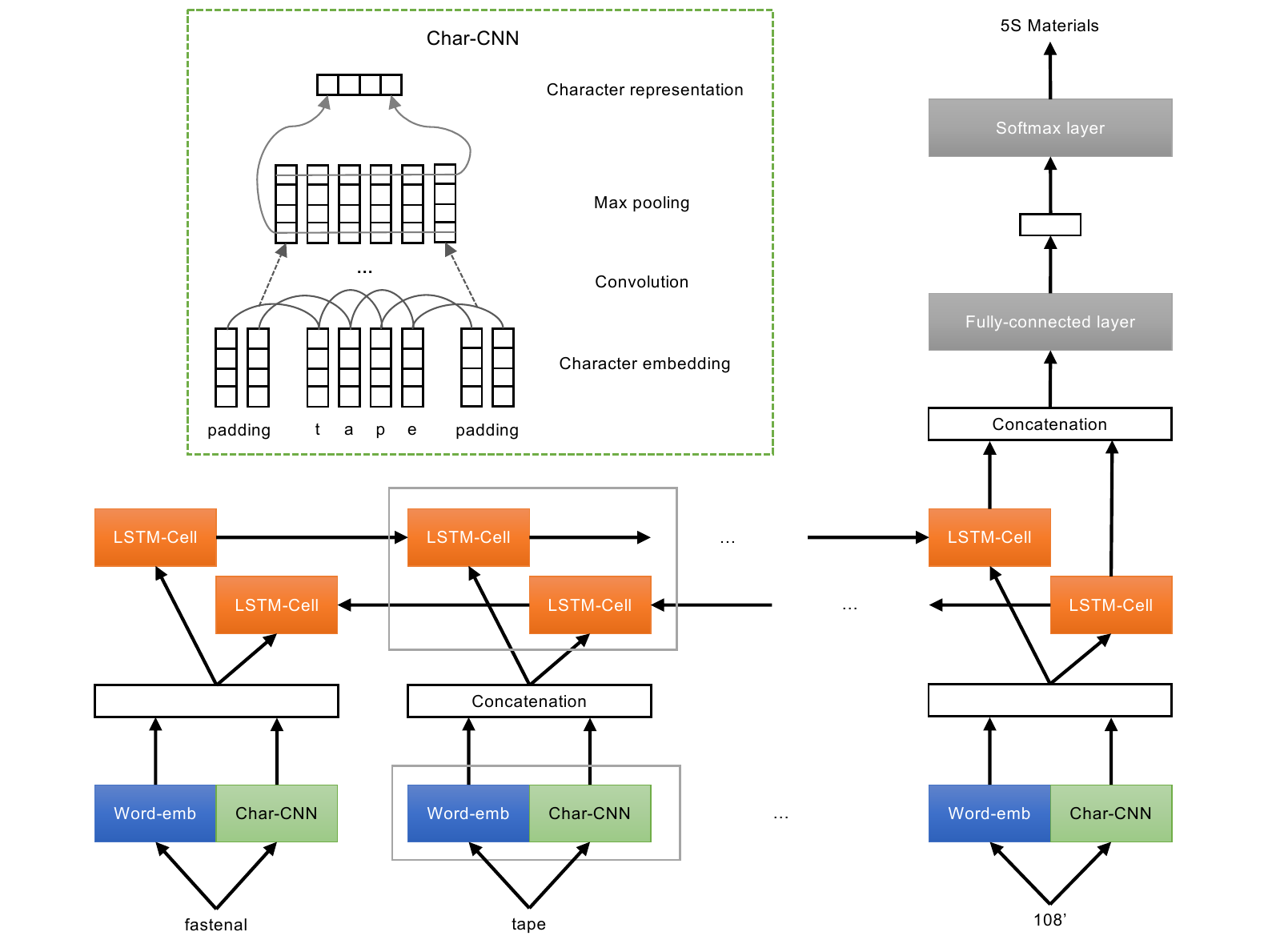}
\end{figure*}

We develop a product title classifier based on LSTM-CNNs architecture proposed in \cite{ma_end--end_2016}. The network architecture is depicted in Figure~\ref{fig:network}. Input encoding layer is a concatenation of word-embeddings (looking-up function against GloVe pre-trained embeddings \cite{pennington_glove_2014}) and character embeddings (output of a Character-CNN layer). The sequence of embedding vectors is passed to a Bidirectional Recurrent Neural Network of LSTM cells \cite{hochreiter_long_1997}. Prediction is carried by a dense layer whose input is last hidden state of Bidirectional LSTM. The DNN is implemented in PyTorch \cite{paszke_automatic_2017} and trained using Adam optimizer with Cross-entropy loss. For experiments with different datasets, we use the same set of hyper-parameters: \emph{Glove embedding} 42B.300d, \emph{LSTM hidden size} 100, \emph{character embedding size} 25 with 3 convolution heads of filter sizes 2, 3, 4, \emph{learning rate} 5e-4, \emph{clip gradient norm} greater than 5.0. Models are trained for 10 \emph{epoch} with \emph{batch size} 16.

To evaluate our implementation, we compare model performance with fine-tunning the pretrained BERT-base uncased language model \cite{devlin_bert_2019}. Results on 6 datasets with clean label are reported in Table~\ref{result-clean}.\footnote{Macro F1 score is a fair evaluation metric for imbalanced data.} Our model performs on par with BERT-base in small datasets Flipkart, WDC, and Retail with macro F1 of less than 1 percentage point lower. For datasets Pricerunner and Skroutz, both models return great performance with BERT-base outperforming our model by 2 percentage points. Shopmania dataset observes the largest performance difference when BERT achieves F1 score 4 percentage points higher than LSTM-CNNs. Good performance of LSTM-CNNs gives us a strong base classifier which is much faster to train than BERT-base (LSTM-CNNs has approximately 6M of trainable parameters while it is 110M for BERT). We will study the impact of pre-training on noise-resistance in a future study.

\begin{table}
  \caption{Models' macro F1 scores on product title data}
  \label{result-clean}
  \centering
  \begin{tabular}{lcc}
    \hline
    Dataset & LSTM-CNNs & BERT-base \\
    \hline
    Flipkart & 0.89  & 0.90 \\
    WDC & 0.92  & 0.92 \\
    Retail & 0.82  & 0.82 \\
    Pricerunner & 0.96  & 0.98 \\
    Shopmania & 0.83 & 0.87 \\
    Skroutz & 0.96  & 0.98 \\
    \hline
  \end{tabular}
\end{table}

\section{Instance-Dependent Noise Stimulation}
\label{sec:idn}

A common approach for automated IDN generation is to train one or a set of classifiers on clean label data, and use such classifiers to generate noisy labels for the whole dataset. Related studies can be different on how to maintain a pool of classifiers, e.g., different checkpoints of a single models or different model architectures, and label placement strategies, e.g., whether replacing clean label samples with noisy counterparts or allowing a sample to have multiple copies with different labels. We follow \cite{zhang_learning_2021, chen_beyond_2021} to use replacement strategy which is considered a more difficult setting. We implement four different IDN algorithms, and adjust parameters to generate noisy label data with noise rates (i.e., ratio of noisy label samples over data size) in two levels: 0.2 (low) and 0.4 (medium).

\smallskip

\noindent \textbf{Last-epoch IDN}: We train a base classifier for 10 epochs to obtain the network corresponding to last epoch checkpoint. The trained network is executed on training data to obtain prediction confidence score (i.e., output of softmax layer) for every sample. Following the formula of noise type-I described in \cite{zhang_learning_2021}, we corrupt item category from the most confident label to the second confident label. This method uses a noise factor parameter to control noise rate, thus we run different trials to probe the noise factors that give us noise rates of interest.

\smallskip

\noindent \textbf{Multi-epoch IDN}: The base classifier is trained for 10 epochs to obtain a sequence of networks corresponding to multiple epoch checkpoints. Each sample is assigned a score as the average of prediction probabilities assigned by network sequences following the algorithm proposed in \cite{chen_beyond_2021}. Potential noisy label should have the highest score among possible labels excluding the ground truth. In particular, data instances are sorted by scores of most likely corrupted labels, and $r$ proportion of top instances will have labels flipped to obtain noise rate $r$.

\smallskip

\noindent \textbf{Multi-model IDN}: Similarly to multi-epoch IDN, we train 5 different versions of the base classifier by varying initial weights to get a network sequence, each network corresponds to last epoch checkpoint (i.e., epoch 10) of a training. Then we apply the same algorithm as in \emph{multi-epoch IDN} to calculate noisy labels.

\smallskip

\noindent \textbf{Similarity-based IDN}: From our experience in product data analyses, we hypothesize that human annotators, and thus machine learning models, may have difficulties in categorizing similar items, e.g., ``\emph{Tara Lifestyle Chhota Bheem Printed Art Plastic Pencil Boxes}'' and ``\emph{Starmark BTS Star Art Polyester Pencil Box}''. Our idea is to locate highly similar items across categories and flip their category labels.

To generate noisy labels, we first calculate textual similarity between items of different categories. We implement two vector-based cosine similarity computations. First, A SentenceTransformer model \cite{reimers_sentence-bert_2019}\footnote{Pretrained model \emph{all-MiniLM-L12-v2}} is used to generate embeddings of product titles. Second, a Tf-Idf model is learned from training set to generate Tf-Idf vectors of input titles. For each pair of product titles, we compare two cosine similarities calculated from sentence embedding vectors and Tf-Idf vectors. The greater score of two methods is assigned as similarity score \texttt{Sim} of two inputs. For each item $i_c$ of category $c$, we record the maximal similarity score \texttt{Maxsim} between it and every item from another category $c'$ of category set $C$:

$\texttt{Maxsim}_{c'}(i_c) = max_j(\texttt{Sim}(i_c, j)) \quad j \in c'$

The sequence of maximal similarity scores of the item is used as weight vector $I_c$ for a multinominal distribution from which we draw a noisy label $\hat{c}$ given the item.

$I_{c} = \{\texttt{Maxsim}_{c'}(i_c) \quad \forall c' \in C, c' \ne c \}$.

$\hat{c} \sim \texttt{Multinomial}(I_c)$

For all items, we assign their $\texttt{Maxsim}_{\hat{c}}$ as representative scores of their corrupted labels, and we sort items by corrupted label scores from high to low. Given noise rate $r$, we select top $r$ proportion of items to replace true labels by corrupted labels.

\section{Experiments on Noisy Labels}

\subsection{Data Denoising by Corrupting Product Titles}
\label{sec:aug}

\begin{table*}
  \caption{Average reduction of noise rate and data size after denoising}
  \label{result-denoise}
  \centering
  \begin{tabular}{lcc|cc}
    \hline
             & \multicolumn{2}{c}{Noise rate 0.2}  & \multicolumn{2}{c}{Noise rate 0.4} \\
    \hline
    Dataset  & Noise reduction & Data reduction & Noise reduction & Data reduction \\
    \hline
    Flipkart & 36\% & 4\% & 29\% & 11\% \\
    WDC & 28\% & 3\% & 21\% & 8\% \\
    Retail & 26\% & 8\% & 30\% & 17\% \\
    Pricerunner & 48\% & 3\% & 43\% & 11\% \\
    Shopmania & 50\% & 7\% & 43\% & 12\% \\
    Skroutz & 44\% & 6\% & 33\% & 7\% \\
    \hline
  \end{tabular}
\end{table*}

We propose a novel data denoising method that reduces noise ratio by relabeling a sample when its prediction is certain. We say an input has certain prediction when model prediction on both original and corrupted inputs are the same. Our method relies on an idea of critical information assumption, i.e., we hypothesize that there are product titles which provide too much information that model does not need to use all words to predict their labels. For such titles, if one or more words are dropped, model should still predict the same label. There have been different studies to extract part of critical information from input to explain output of prediction models \cite{ribeiro_why_2016, lundberg_unified_2017, kokalj_bert_2021}. Regarding product title, leading words are considerately more important than trailing words for recognizing product category.\footnote{A common template arranges title words in order of Brand Name > Product > Key features > Size > Color > Quantity (sellerengine.com/product-title-keyword-strategies-for-new-products-on-amazon).} Algorithm~\ref{algo:drop} is a simple heuristic to drop words from a product title. Statement 2 makes sure some right words are dropped even when an input is less than 15 words.

\begin{algorithm}
\caption{Drop words from a product title}\label{algo:drop}
\begin{algorithmic}[1]
\State Drop left words until dropped words have at least 5 letters in total or less then 4 words remaining
\State Drop right words until dropped words have at least 5 letters in total or less then 4 words remaining
\State Drop right words while there are more than 15 words
\end{algorithmic}
\end{algorithm}

We propose Algorithm~\ref{algo:aug} to denoise training data. With clean data, model should achieves highly confident predictions on training samples. Thus, we reason that unconfident predictions on training samples (i.e., $p \le 0.8$) are likely due to noisy labels. We note that in case of noisy training, input label is not considered ground truth generally.

Steps 3 and 4 update\footnote{For efficiency, our actual implementation only update a sample when its input label is different from predicted label. This condition is ignored in pseudo code for simplicity.} training samples while step 5 removes samples which the model is unsure. Our denoising algorithm reduces noise rate with a trade-off of smaller training data. Their impact to training data is shown in Table~\ref{result-denoise}. For each dataset and input noise rate, we average noise rate and data size reductions after denoising the data corrputed by different noise stimulations.

\begin{algorithm}
\caption{Denoise training data}\label{algo:aug}
\begin{algorithmic}[1]
\State Run pre-trained model $M$ on training data $D$: $\{L_o, P_o\} \gets M(D)$ where $L_o$ are predicted label and $P_o$ are prediction probability
\State Run $M$ on corrupted training data $\hat{D}$ (i.e., drops words from titles): $\{L_d, P_d\} \gets M(\hat{D})$
\State Assign predicted labels to samples where predictions are confident:

$\textrm{InputLabel} \gets L_o \textbf{ if } P_o \geq$ 0.8

\State Assign predicted labels to samples where predictions are certain:

$\textrm{InputLabel} \gets L_o \textbf{ if } L_o = L_d$

%\State Remove samples where predictions are neither certain nor correct: $L_o \neq L_d \textbf{ and } L_o \neq \textrm{InputLabel} \textbf{ and } L_d \neq \textrm{InputLabel}$

\State Remove samples where predictions are neither certain nor confident: $L_o \neq L_d \textbf{ and } P_o \le 0.8 \textbf{ and } P_d \le 0.8$
\end{algorithmic}
\end{algorithm}

\begin{table*}
  \caption{Models' macro F1 scores on product title data with noisy labels. Highest scores are bold. API shows average performance improvement compared to base classifier.}
  \label{result-noisy}
  \centering
  \begin{tabular}{lccccc|ccccc}
    \hline
             & \multicolumn{5}{c}{Noise rate 0.2}  & \multicolumn{5}{c}{Noise rate 0.4} \\
             & \multicolumn{10}{c}{\textbf{Last-epoch IDN}} \\
    \hline
    \textbf{Dataset}  & \textbf{Base} & \textbf{DeN} & \textbf{SEAL} & \textbf{PLC} & \textbf{CTp}  & \textbf{Base} & \textbf{DeN} & \textbf{SEAL} & \textbf{PLC} & \textbf{CTp} \\
    \hline
    Flipkart & 0.74 & \textbf{0.82} & 0.81 & 0.78 & 0.81 & 0.55 & 0.67 & \textbf{0.69} & 0.62 & 0.66 \\
    WDC & 0.86 & 0.86 & \textbf{0.88} & 0.87 & \textbf{0.88} & 0.68 & 0.71 & 0.71 & 0.73 & \textbf{0.77} \\
    Retail & 0.72 & 0.76 & \textbf{0.78} & \textbf{0.78} & \textbf{0.78} & 0.59 & 0.66 & 0.70 & 0.66 & \textbf{0.71} \\
    Pricerunner & 0.89 & \textbf{0.94}  & \textbf{0.94} & 0.93 & \textbf{0.94} & 0.71 & 0.87 & 0.90 & 0.79 & \textbf{0.91} \\
    Shopmania & 0.74 & 0.71 & 0.73 & \textbf{0.76} & 0.68 & 0.59 & 0.62 & 0.62 & \textbf{0.63} & 0.56 \\
    Skroutz & 0.90 & 0.94 & 0.94 & 0.93 & \textbf{0.95} & 0.77 & 0.86 & 0.86 & 0.78 & \textbf{0.92} \\
    \hline
    \emph{API} & - & 3.7\% & 4.8\% & 4.2\% & 3.8\% & - & 12.9\% & 15.3\% & 8.5\% & 16\% \\
    \hline
             & \multicolumn{10}{c}{\textbf{Multi-epoch IDN}} \\
    \hline
    Flipkart & 0.73 & 0.73  & 0.74 & \textbf{0.75} & \textbf{0.75} & 0.61 & 0.59 & \textbf{0.64} & 0.62 & 0.63 \\
    WDC & 0.81 & 0.82 & \textbf{0.83} & \textbf{0.83} & 0.82 & 0.65 & 0.66 & 0.66 & 0.65 & \textbf{0.68} \\
    Retail & 0.79 & \textbf{0.80} & 0.79 & 0.79 & \textbf{0.80} & 0.73 & 0.73 & \textbf{0.76} & 0.74 & \textbf{0.76} \\
    Pricerunner & 0.91 & 0.91  & \textbf{0.92} & \textbf{0.92} & \textbf{0.92} & 0.80 & 0.82 & 0.84 & 0.82 & \textbf{0.85} \\
    Shopmania &  0.76 & 0.75 & 0.76 & \textbf{0.77} & 0.67 & 0.63 & \textbf{0.65} & \textbf{0.65} & 0.62 & 0.57 \\
    Skroutz &  \textbf{0.95} & \textbf{0.95} & \textbf{0.95} & \textbf{0.95} & \textbf{0.95} & 0.88 & \textbf{0.90} & \textbf{0.90} & 0.88 & \textbf{0.90} \\
    \hline
    \emph{API} & - & 0.2\% & 0.8\% & 1.3\% & -0.9\% & - & 1\% & 3.5\% & 0.6\% & 1.8\% \\
    \hline
             & \multicolumn{10}{c}{\textbf{Multi-model IDN}} \\
    \hline
    Flipkart & 0.72 & 0.74  & \textbf{0.75} & 0.74 & \textbf{0.75} & 0.57 & 0.61 & \textbf{0.64} & 0.61 & 0.63 \\
    WDC & 0.82 & \textbf{0.83}  & \textbf{0.83} & 0.82 & \textbf{0.83} & 0.65 & 0.65 & \textbf{0.67} & 0.66 & \textbf{0.67} \\
    Retail & 0.78 & 0.79 & \textbf{0.80} & 0.79 & 0.79 & 0.70 & 0.73 & \textbf{0.76} & 0.73 & 0.74 \\
    Pricerunner & 0.90 & 0.91  & \textbf{0.92} & 0.91 & \textbf{0.92} & 0.80 & 0.81 & \textbf{0.84} & 0.81 & \textbf{0.84} \\
    Shopmania &  0.76 & 0.75 & \textbf{0.78} & 0.77 & 0.68 & \textbf{0.66} & 0.65 & \textbf{0.66} & 0.64 & 0.57 \\
    Skroutz & \textbf{0.95} & \textbf{0.95} & \textbf{0.95} & \textbf{0.95} & \textbf{0.95} & 0.90 & \textbf{0.92} & 0.91 & 0.91 & \textbf{0.92} \\
    \hline
    \emph{API} & - & 0.8\% & 2.1\% & 1\% & -0.2\% & - & 2.2\% & 5\% & 2\% & 2.1\% \\
    \hline
             & \multicolumn{10}{c}{\textbf{Similarity-based IDN}} \\
    \hline
    Flipkart & 0.73 & 0.76  & 0.76 & 0.77 & \textbf{0.78} & 0.55 & 0.58 & 0.61 & 0.65 & \textbf{0.67} \\
    WDC & 0.73 & 0.74  & 0.75 & 0.75 & \textbf{0.76} & 0.58 & 0.58 & 0.59 & 0.59 & \textbf{0.60} \\
    Retail & 0.69 & 0.75 & \textbf{0.77} & 0.76 & \textbf{0.77} & 0.57 & 0.66 & \textbf{0.72} & 0.70 & \textbf{0.72} \\
    Pricerunner & 0.86 & 0.91  & \textbf{0.93} & 0.92 & \textbf{0.93} & 0.72 & 0.83 & 0.85 & 0.82 & \textbf{0.86} \\
    Shopmania & 0.70 & 0.70 & 0.71 & \textbf{0.73} & 0.65 & 0.57 & 0.59 & 0.57 & \textbf{0.59} & 0.50 \\
    Skroutz & 0.84 & \textbf{0.89} & 0.85 & 0.84 & 0.88 & 0.68 & \textbf{0.76} & 0.72 & 0.69 & \textbf{0.76} \\
    \hline
    \emph{API} & - & 4.3\% & 4.8\% & 4.9\% & 4.7\% & - & 8.6\% & 10.4\% & 10.2\% & 11.7\% \\
    \hline
  \end{tabular}
\end{table*}

\begin{table*}
  \caption{Models' macro F1 scores averaged over different noise stimulations. Highest scores are bold.}
  \label{result-noisy2}
  \centering
  \begin{tabular}{llllll|lllll}
    \hline
             & \multicolumn{5}{c}{Noise rate 0.2}  & \multicolumn{5}{c}{Noise rate 0.4} \\
    \hline
    \textbf{Dataset}  & \textbf{Base} & \textbf{DeN} & \textbf{SEAL} & \textbf{PLC} & \textbf{CTp}  & \textbf{Base} & \textbf{DeN} & \textbf{SEAL} & \textbf{PLC} & \textbf{CTp} \\
    \hline
    Flipkart & 0.73 & 0.762 & 0.765 & 0.76 & \textbf{0.772} & 0.57 & 0.6125 & 0.645 & 0.625 & \textbf{0.647} \\
    WDC & 0.805 & 0.812 & 0.822 & 0.817 & \textbf{0.822} & 0.64 & 0.65 & 0.6575 & 0.657 & \textbf{0.68} \\
    Retail & 0.745 & 0.775 & \textbf{0.785} & 0.78 & \textbf{0.785} & 0.6475 & 0.695 & \textbf{0.735} & 0.707 & 0.732 \\
    Pricerunner & 0.89 & 0.917 & \textbf{0.927} & 0.92 & \textbf{0.927} & 0.757 & 0.832 & 0.857 & 0.81 & \textbf{0.865} \\
    Shopmania & 0.74 & 0.727 & 0.745 & \textbf{0.757} & 0.67 & 0.612 & \textbf{0.627} & 0.625 & 0.62 & 0.55 \\
    Skroutz & 0.91 & \textbf{0.932} & 0.9225 & 0.917 & \textbf{0.932} & 0.807 & 0.86 & 0.8475 & 0.815 & \textbf{0.875} \\
    \hline
  \end{tabular}
\end{table*}

\subsection{Noise-Resistance Training Algorithms}
\label{sec:train}

In this study, we compare three training solutions that were developed for data with noisy labels: Self-Evolution Average Label -- SEAL~\cite{chen_beyond_2021}, Progressive Label Correction -- PLC~\cite{zhang_learning_2021} and CoTeaching$^+$ -- CTp \cite{yu_how_2019}. The three training algorithms work independently from the underlying models.

SEAL trains a model on multiple iterations. In each iteration, SEAL optimizes model's loss against soft labels which are average predictions over epochs of the previous iteration. PLC first trains noisy label data normally for a number of epochs, i.e., warm-up phase, with expectation that model can learn from clean labels before over-fits to noisy labels. Then PLC corrects input labels after each epoch for cases that it yields a confidence score above a threshold. CoTeaching$^+$ is an upgrade of CoTeaching paradigm that cross-trains two models using only small-loss samples in each mini-batch. CoTeaching$^+$ further prevents the two models from convergence by passing only samples whose predictions disagree among small-loss data to loss optimization step.

\subsection{Experiment Results}
\label{sec:result}

Experimental results of individual models are shown in Table~\ref{result-noisy}. We first train the base classifier directly on noisy label data and record Macro F1 score on column \emph{Base}. We then denoise\footnote{We run pre-trained model reported in column Base on training data to collect prediction outputs as described in Algorithm~\ref{algo:aug}.} training data before training the base classifier, and enter performance into column \emph{DeN}. Next columns report F1 scores of models trained by noise-resistance algorithms on noisy label data (i.e., not desnoised).

As expected, label noises degrade model performance significantly. Noise rate 0.2 reduces performance of base model from 5\% (Skroutz) - 18\% (Flipkart), while the performance reduction is 17\% (Skroutz) to 46\% (Flipkart) given noise rate 0.4. Pricerunner and Skroutz have lowest performance degradation which is reasonable because these two datasets are the easiest (see Table~\ref{result-clean}).

Evaluating impact of different IDN methods, similarity-based IDN degrades performance of base classifier the most in comparison with other IDN methods. Comparing performance of noise-resistance training methods with base classifier, we report \textbf{a}verage \textbf{p}erformance \textbf{i}mprovement (API) over different datasets in percentage point. Noise-resistance training methods have the most difficulty in improving multi-epoch and multi-model IDNs. In particular, performance improvements are at most 2\% and 5\% when multi-epoch and multi-model IDN rates are 0.2 and 0.4 respectively. Such noise-resistance training methods achieve much higher performance improvements when noisy labels are generated by other two IDN methods. Particularly, average performance improves are at least 4\% and 8\% when last-epoch and similarity-based IDN rates are 0.2 and 0.4 respectively.

Denoising data before training show improvements but performance improvements are lower for multi-epoch and multi-model IDN's than for last-epoch and similarity-based IDN's. Although our data denoising implementation is basic, it helps improve performance more than PLC in many settings, e.g., higher API in last-epoch, multi-epoch and multi-model IDN's. This encourage us to explore more advanced classifiers for better noise reduction results.

Table~\ref{result-noisy2} summarizes the results by grouping by dataset name then averaging over different noise stimulation methods. It is shown that CoTeaching$^+$ performs better than other methods in many datasets, e.g., 5 datasets with noise rate 0.2 and 4 datasets out of 6 with noise rate 0.4. DeN performs worse than three noise-resistance training methods despite a fact that noise rate was reduced significantly as shown in Table~\ref{result-denoise}. We hypothesize that regular training cannot recover from noisy instances that denoising algorithm is unable to correct/remove.

Comparing different datasets, we observe that Shopmania is the most difficult. Among denoising and noise-resistance training algorithms, the best approach could only improve performance by 4\% and 7\% when noise rate is 0.2 and 0.4 respectively. CoTeaching$^+$ even performed worse than base classifier on this dataset. As shown in Table~\ref{data}, Shopmania is the largest dataset, has the most number of classes and the most imbalanced distribution. Regarding imbalanced data, noisy labels in a minor class might be harder to address due to its small number of instances.

Finally, prediction performance at high noise rate 0.6 is briefly shown in Table~\ref{result-noisy6}. We only compare base classifier to CoTeaching$^+$ which is the best performing approach in this setting. While noise-resistance training algorithms do improve performance, overall performance is low. In our opinion, such a performance score is too low for an product title classification application. Thus we do not find any of the three training algorithms or our denoising algorithm can work reasonably well with high noise rate in product data.

\begin{table}
  \caption{Models' macro F1 scores on product title data with noise rate 0.6. Scores are averaged over IDN methods.}
  \label{result-noisy6}
  \centering
  \begin{tabular}{lccc}
    \hline
    Dataset & Base & CTp & API (\%) \\
    \hline
    Flipkart & 0.41  & 0.45 & 10\% \\
    WDC & 0.42  & 0.46 & 9\% \\
    Retail & 0.52 & 0.60 & 15\% \\
    Pricerunner & 0.51 & 0.54 & 6\% \\
    Shopmania & 0.42 & 0.42 & 0\% \\
    Skroutz & 0.58 & 0.64 & 10\% \\
    \hline
  \end{tabular}
\end{table}

\subsection{Future Work}

Data denoising algorithm opens new opportunities for us to further improve product title classification with noisy labels. We plan to improve data denoising by several techniques: (1) run denoising algorithm using a base model trained with small number of epochs to prevent over-fitting to noise, (2) use more advanced base classifier, and transformer-based model is a good candidate. Stacking data denoising and noise-resistance training is another extension, and we can approach this in two ways: (1) data denoising provides less-noisy data for noise-resistance training, (2) noise-resistance training provides better base model to denoise data.

\section{Conclusion}

In this paper, we evaluate a denoising algorithm and three training approaches for product title classification with category labels corrupted by instance-dependent noise. We introduce a new IDN stimulation algorithm and compare with three IDN algorithms from prior studies to explore model performance on a wider range of noise type. Therefore our study can evaluate model robustness to IDN more reliably. Overall we find that CoTeaching$^+$ achieves highest average improvement and be our recommendation when applying to new product data without prior knowledge of noise cause or true distribution. SEAL can be a good method when we have clean validation data to evaluate. However, all methods studied in this paper have difficulties to address noise in large scale data with highly imbalanced class distribution, especially when noise rate is high. For such extreme setting, application of data denoising and noise-resistance training algorithms could not yield to reasonable performance for applying to production. For a future work, we plan to combine multiple techniques including transformer-based classifier as a more advanced model and stacking data denoising with noise-resistance training.

\bibliography{product}

\begin{thebibliography}{32}
\expandafter\ifx\csname natexlab\endcsname\relax\def\natexlab#1{#1}\fi

\bibitem[{Akritidis et~al.(2018)Akritidis, Fevgas, and
  Bozanis}]{akritidis_effective_2018}
Leonidas Akritidis, Athanasios Fevgas, and Panayiotis Bozanis. 2018.
\newblock \href {https://doi.org/10.1109/ICTAI.2018.00041} {Effective
  {Products} {Categorization} with {Importance} {Scores} and {Morphological}
  {Analysis} of the {Titles}}.
\newblock In \emph{2018 {IEEE} 30th {International} {Conference} on {Tools}
  with {Artificial} {Intelligence} ({ICTAI})}, pages 213--220.

\bibitem[{Akritidis et~al.(2020)Akritidis, Fevgas, Bozanis, and
  Makris}]{akritidis_self-verifying_2020}
Leonidas Akritidis, Athanasios Fevgas, Panayiotis Bozanis, and Christos Makris.
  2020.
\newblock A self-verifying clustering approach to unsupervised matching of
  product titles.
\newblock \emph{Artificial Intelligence Review}, pages 1--44.

\bibitem[{Arpit et~al.(2017)Arpit, Jastrzundefinedbski, Ballas, Krueger,
  Bengio, Kanwal, Maharaj, Fischer, Courville, Bengio, and
  Lacoste-Julien}]{arpit_closer_2017}
Devansh Arpit, Stanis{\textbackslash}law Jastrzundefinedbski, Nicolas Ballas,
  David Krueger, Emmanuel Bengio, Maxinder~S. Kanwal, Tegan Maharaj, Asja
  Fischer, Aaron Courville, Yoshua Bengio, and Simon Lacoste-Julien. 2017.
\newblock A {Closer} {Look} at {Memorization} in {Deep} {Networks}.
\newblock In \emph{Proceedings of the 34th {International} {Conference} on
  {Machine} {Learning} - {Volume} 70}, {ICML}'17, pages 233--242. JMLR.org.
\newblock Event-place: Sydney, NSW, Australia.

\bibitem[{Brinkmann and Bizer(2021)}]{brinkmann_improving_2021}
Alexander Brinkmann and Christian Bizer. 2021.
\newblock Improving hierarchical product classification using domain-specific
  language modelling.
\newblock In \emph{Proceedings of {Workshop} on {Knowledge} {Management} in
  e-{Commerce}}.

\bibitem[{Chen et~al.(2021{\natexlab{a}})Chen, Chou, Xia, and
  Miyake}]{chen_multimodal_2021}
Lei Chen, Houwei Chou, Yandi Xia, and Hirokazu Miyake. 2021{\natexlab{a}}.
\newblock \href {https://doi.org/10.18653/v1/2021.ecnlp-1.13} {Multimodal
  {Item} {Categorization} {Fully} {Based} on {Transformer}}.
\newblock In \emph{Proceedings of {The} 4th {Workshop} on e-{Commerce} and
  {NLP}}, pages 111--115, Online. Association for Computational Linguistics.

\bibitem[{Chen et~al.(2021{\natexlab{b}})Chen, Ye, Chen, Zhao, and
  Heng}]{chen_beyond_2021}
Pengfei Chen, Junjie Ye, Guangyong Chen, Jingwei Zhao, and Pheng-Ann Heng.
  2021{\natexlab{b}}.
\newblock Beyond {Class}-{Conditional} {Assumption}: {A} {Primary} {Attempt} to
  {Combat} {Instance}-{Dependent} {Label} {Noise}.
\newblock In \emph{Proceedings of the {AAAI} {Conference} on {Artificial}
  {Intelligence}}.

\bibitem[{Das et~al.(2016)Das, Xia, Levine, Di~Fabbrizio, and
  Datta}]{das_large-scale_2016}
Pradipto Das, Yandi Xia, Aaron Levine, Giuseppe Di~Fabbrizio, and Ankur Datta.
  2016.
\newblock \href {https://doi.org/10.1109/BigData.2016.7841063} {Large-scale
  taxonomy categorization for noisy product listings}.
\newblock In \emph{2016 {IEEE} {International} {Conference} on {Big} {Data}
  ({Big} {Data})}, pages 3885--3894.

\bibitem[{Devlin et~al.(2019)Devlin, Chang, Lee, and
  Toutanova}]{devlin_bert_2019}
Jacob Devlin, Ming-Wei Chang, Kenton Lee, and Kristina Toutanova. 2019.
\newblock \href {https://doi.org/10.18653/v1/N19-1423} {{BERT}: {Pre}-training
  of {Deep} {Bidirectional} {Transformers} for {Language} {Understanding}}.
\newblock In \emph{Proceedings of the 2019 {Conference} of the {North}
  {American} {Chapter} of the {Association} for {Computational} {Linguistics}:
  {Human} {Language} {Technologies}, {Volume} 1 ({Long} and {Short} {Papers})},
  pages 4171--4186, Minneapolis, Minnesota. Association for Computational
  Linguistics.

\bibitem[{Elayanithottathil and Keuper(2021)}]{elayanithottathil_retail_2021}
Febin~Sebastian Elayanithottathil and Janis Keuper. 2021.
\newblock A {Retail} {Product} {Categorisation} {Dataset}.
\newblock \_eprint: 2103.13864.

\bibitem[{Gao et~al.(2020)Gao, Yang, Zhou, Wei, Hu, and Wang}]{gao_deep_2020}
Dehong Gao, Wenjing Yang, Huiling Zhou, Yi~Wei, Y.~Hu, and H.~Wang. 2020.
\newblock Deep {Hierarchical} {Classification} for {Category} {Prediction} in
  {E}-commerce {System}.
\newblock \emph{ArXiv}, abs/2005.06692.

\bibitem[{Garg et~al.(2021)Garg, Ramakrishnan, and Thumbe}]{garg_towards_2021}
Siddhant Garg, Goutham Ramakrishnan, and Varun Thumbe. 2021.
\newblock Towards {Robustness} to {Label} {Noise} in {Text} {Classification}
  via {Noise} {Modeling}.
\newblock \emph{Proceedings of the 30th ACM International Conference on
  Information \& Knowledge Management}.

\bibitem[{Gu et~al.(2021)Gu, Masotto, Bachani, Lakshminarayanan, Nikodem, and
  Yin}]{gu_realistic_2021}
Keren Gu, Xander Masotto, Vandana Bachani, Balaji Lakshminarayanan, Jack
  Nikodem, and Dong Yin. 2021.
\newblock A {Realistic} {Simulation} {Framework} for {Learning} with {Label}
  {Noise}.
\newblock \emph{ArXiv}, abs/2107.11413.

\bibitem[{Han et~al.(2018)Han, Yao, Yu, Niu, Xu, Hu, Tsang, and
  Sugiyama}]{han_co-teaching_2018}
Bo~Han, Quanming Yao, Xingrui Yu, Gang Niu, Miao Xu, Weihua Hu, Ivor Tsang, and
  Masashi Sugiyama. 2018.
\newblock Co-teaching: {Robust} training of deep neural networks with extremely
  noisy labels.
\newblock \emph{Advances in neural information processing systems}, 31.

\bibitem[{Hochreiter and Schmidhuber(1997)}]{hochreiter_long_1997}
Sepp Hochreiter and Jürgen Schmidhuber. 1997.
\newblock \href {https://doi.org/10.1162/neco.1997.9.8.1735} {Long
  {Short}-{Term} {Memory}}.
\newblock \emph{Neural Comput.}, 9(8):1735--1780.
\newblock Place: Cambridge, MA, USA Publisher: MIT Press.

\bibitem[{Jiang et~al.(2018)Jiang, Zhou, Leung, Li, and
  Fei-Fei}]{jiang_mentornet_2018}
Lu~Jiang, Zhengyuan Zhou, Thomas Leung, Li-Jia Li, and Li~Fei-Fei. 2018.
\newblock {MentorNet}: {Learning} {Data}-{Driven} {Curriculum} for {Very}
  {Deep} {Neural} {Networks} on {Corrupted} {Labels}.
\newblock In \emph{{ICML}}.

\bibitem[{Jindal et~al.(2016)Jindal, Nokleby, and Chen}]{jindal_learning_2016}
Ishan Jindal, Matthew Nokleby, and Xuewen Chen. 2016.
\newblock Learning deep networks from noisy labels with dropout regularization.
\newblock In \emph{Data {Mining} ({ICDM}), 2016 {IEEE} 16th {International}
  {Conference} on}, pages 967--972. IEEE.

\bibitem[{Jindal et~al.(2019)Jindal, Pressel, Lester, and
  Nokleby}]{jindal_effective_2019}
Ishan Jindal, Daniel Pressel, Brian Lester, and Matthew Nokleby. 2019.
\newblock \href {https://doi.org/10.18653/v1/N19-1328} {An {Effective} {Label}
  {Noise} {Model} for {DNN} {Text} {Classification}}.
\newblock In \emph{Proceedings of the 2019 {Conference} of the {North}
  {American} {Chapter} of the {Association} for {Computational} {Linguistics}:
  {Human} {Language} {Technologies}, {Volume} 1 ({Long} and {Short} {Papers})},
  pages 3246--3256, Minneapolis, Minnesota. Association for Computational
  Linguistics.

\bibitem[{Kokalj et~al.(2021)Kokalj, Škrlj, Lavrač, Pollak, and
  Robnik-Šikonja}]{kokalj_bert_2021}
Enja Kokalj, Blaž Škrlj, Nada Lavrač, Senja Pollak, and Marko
  Robnik-Šikonja. 2021.
\newblock \href {https://aclanthology.org/2021.hackashop-1.3} {{BERT} meets
  {Shapley}: {Extending} {SHAP} {Explanations} to {Transformer}-based
  {Classifiers}}.
\newblock In \emph{Proceedings of the {EACL} {Hackashop} on {News} {Media}
  {Content} {Analysis} and {Automated} {Report} {Generation}}, pages 16--21,
  Online. Association for Computational Linguistics.

\bibitem[{Lundberg and Lee(2017)}]{lundberg_unified_2017}
Scott~M. Lundberg and Su-In Lee. 2017.
\newblock A {Unified} {Approach} to {Interpreting} {Model} {Predictions}.
\newblock In \emph{Proceedings of the 31st {International} {Conference} on
  {Neural} {Information} {Processing} {Systems}}, {NIPS}'17, pages 4768--4777,
  Red Hook, NY, USA. Curran Associates Inc.
\newblock Event-place: Long Beach, California, USA.

\bibitem[{Ma and Hovy(2016)}]{ma_end--end_2016}
Xuezhe Ma and Eduard Hovy. 2016.
\newblock \href {https://doi.org/10.18653/v1/P16-1101} {End-to-end {Sequence}
  {Labeling} via {Bi}-directional {LSTM}-{CNNs}-{CRF}}.
\newblock In \emph{Proceedings of the 54th {Annual} {Meeting} of the
  {Association} for {Computational} {Linguistics} ({Volume} 1: {Long}
  {Papers})}, pages 1064--1074, Berlin, Germany. Association for Computational
  Linguistics.

\bibitem[{Malach and Shalev-Shwartz(2017)}]{malach_decoupling_2017}
Eran Malach and Shai Shalev-Shwartz. 2017.
\newblock Decoupling "when to update" from "how to update".
\newblock In \emph{{NIPS}}.

\bibitem[{Paszke et~al.(2017)Paszke, Gross, Chintala, Chanan, Yang, DeVito,
  Lin, Desmaison, Antiga, and Lerer}]{paszke_automatic_2017}
Adam Paszke, Sam Gross, Soumith Chintala, Gregory Chanan, Edward Yang, Zachary
  DeVito, Zeming Lin, Alban Desmaison, Luca Antiga, and Adam Lerer. 2017.
\newblock Automatic differentiation in {PyTorch}.
\newblock In \emph{{NIPS}-{W}}.

\bibitem[{Patrini et~al.(2017)Patrini, Rozza, Krishna~Menon, Nock, and
  Qu}]{patrini_making_2017}
Giorgio Patrini, Alessandro Rozza, Aditya Krishna~Menon, Richard Nock, and
  Lizhen Qu. 2017.
\newblock Making deep neural networks robust to label noise: {A} loss
  correction approach.
\newblock In \emph{Proceedings of the {IEEE} conference on computer vision and
  pattern recognition}, pages 1944--1952.

\bibitem[{Pennington et~al.(2014)Pennington, Socher, and
  Manning}]{pennington_glove_2014}
Jeffrey Pennington, Richard Socher, and Christopher~D. Manning. 2014.
\newblock \href {http://www.aclweb.org/anthology/D14-1162} {{GloVe}: {Global}
  {Vectors} for {Word} {Representation}}.
\newblock In \emph{Empirical {Methods} in {Natural} {Language} {Processing}
  ({EMNLP})}, pages 1532--1543.

\bibitem[{Primpeli et~al.(2019)Primpeli, Peeters, and
  Bizer}]{primpeli_wdc_2019}
Anna Primpeli, Ralph Peeters, and Christian Bizer. 2019.
\newblock \href {https://doi.org/10.1145/3308560.3316609} {The {WDC} {Training}
  {Dataset} and {Gold} {Standard} for {Large}-{Scale} {Product} {Matching}}.
\newblock In \emph{Companion {Proceedings} of {The} 2019 {World} {Wide} {Web}
  {Conference}}, {WWW} '19, pages 381--386, New York, NY, USA. Association for
  Computing Machinery.
\newblock Event-place: San Francisco, USA.

\bibitem[{Reimers and Gurevych(2019)}]{reimers_sentence-bert_2019}
Nils Reimers and Iryna Gurevych. 2019.
\newblock \href {https://arxiv.org/abs/1908.10084} {Sentence-{BERT}: {Sentence}
  {Embeddings} using {Siamese} {BERT}-{Networks}}.
\newblock In \emph{Proceedings of the 2019 {Conference} on {Empirical}
  {Methods} in {Natural} {Language} {Processing}}. Association for
  Computational Linguistics.

\bibitem[{Ribeiro et~al.(2016)Ribeiro, Singh, and Guestrin}]{ribeiro_why_2016}
Marco~Tulio Ribeiro, Sameer Singh, and Carlos Guestrin. 2016.
\newblock "{Why} {Should} {I} {Trust} {You}?": {Explaining} the {Predictions}
  of {Any} {Classifier}.
\newblock In \emph{Proceedings of the 22nd {ACM} {SIGKDD} {International}
  {Conference} on {Knowledge} {Discovery} and {Data} {Mining}, {San}
  {Francisco}, {CA}, {USA}, {August} 13-17, 2016}, pages 1135--1144.

\bibitem[{Shen et~al.(2012)Shen, Ruvini, and Sarwar}]{shen_large-scale_2012}
Dan Shen, Jean-David Ruvini, and Badrul Sarwar. 2012.
\newblock \href {https://doi.org/10.1145/2396761.2396838} {Large-{Scale} {Item}
  {Categorization} for e-{Commerce}}.
\newblock In \emph{Proceedings of the 21st {ACM} {International} {Conference}
  on {Information} and {Knowledge} {Management}}, {CIKM} '12, pages 595--604,
  New York, NY, USA. Association for Computing Machinery.
\newblock Event-place: Maui, Hawaii, USA.

\bibitem[{Song et~al.(2022)Song, Kim, Park, Shin, and Lee}]{song_learning_2022}
Hwanjun Song, Minseok Kim, Dongmin Park, Yooju Shin, and Jae-Gil Lee. 2022.
\newblock Learning from {Noisy} {Labels} with {Deep} {Neural} {Networks}: {A}
  {Survey}.
\newblock \emph{IEEE Transactions on Neural Networks and Learning Systems}.

\bibitem[{Yu et~al.(2019)Yu, Han, Yao, Niu, Tsang, and Sugiyama}]{yu_how_2019}
Xingrui Yu, Bo~Han, Jiangchao Yao, Gang Niu, Ivor Tsang, and Masashi Sugiyama.
  2019.
\newblock How does disagreement help generalization against label corruption?
\newblock In \emph{International {Conference} on {Machine} {Learning}}, pages
  7164--7173. PMLR.

\bibitem[{Zhang et~al.(2021{\natexlab{a}})Zhang, Bengio, Hardt, Recht, and
  Vinyals}]{zhang_understanding_2021}
Chiyuan Zhang, Samy Bengio, Moritz Hardt, Benjamin Recht, and Oriol Vinyals.
  2021{\natexlab{a}}.
\newblock \href {https://doi.org/10.1145/3446776} {Understanding {Deep}
  {Learning} ({Still}) {Requires} {Rethinking} {Generalization}}.
\newblock \emph{Commun. ACM}, 64(3):107--115.
\newblock Place: New York, NY, USA Publisher: Association for Computing
  Machinery.

\bibitem[{Zhang et~al.(2021{\natexlab{b}})Zhang, Zheng, Wu, Goswami, and
  Chen}]{zhang_learning_2021}
Yikai Zhang, Songzhu Zheng, Pengxiang Wu, Mayank Goswami, and Chao Chen.
  2021{\natexlab{b}}.
\newblock Learning with {Feature}-{Dependent} {Label} {Noise}: {A}
  {Progressive} {Approach}.
\newblock In \emph{{ICLR}}.

\end{thebibliography}

\end{document}